\documentclass[twoside]{article}

\usepackage[accepted]{aistats2019}

\usepackage{bm,color,soul}
\usepackage{amsmath,amsfonts,amssymb}
\usepackage{mathtools}
\usepackage[retainorgcmds]{IEEEtrantools}
\usepackage{boldline}
\usepackage{setspace}

\usepackage[round]{natbib}

\begin{document}

%

%

\runningauthor{Ardywibowo, Zhao, Wang, Mortazavi, Huang, Qian}

\twocolumn[

\aistatstitle{Adaptive Activity Monitoring with Uncertainty Quantification in Switching Gaussian Process Models}

\aistatsauthor{Randy Ardywibowo$^{1}$ \And Guang Zhao$^{1}$ \And Zhangyang Wang$^{1}$ \And  Bobak Mortazavi$^{1}$}
\onehalfspacing

\aistatsauthor{Shuai Huang$^{2}$ \And  Xiaoning Qian$^{1}$  }

\aistatsaddress{ Texas A\&M University$^{1}$ \And University of Washington$^{2}$ } 

]

\begin{abstract}
Emerging wearable sensors have enabled the unprecedented ability to continuously monitor human activities for healthcare purposes. However, with so many ambient sensors collecting different measurements, it becomes important not only to maintain good monitoring accuracy, but also low power consumption to ensure sustainable monitoring. This power-efficient sensing scheme can be achieved by deciding which group of sensors to use at a given time, requiring an accurate characterization of the trade-off between sensor energy usage and the uncertainty in ignoring certain sensor signals while monitoring. To address this challenge in the context of activity monitoring, we have designed an adaptive activity monitoring framework. We first propose a \textit{switching Gaussian process} to model the observed sensor signals emitting from the underlying activity states. To efficiently compute the Gaussian process model likelihood and quantify the context prediction uncertainty, we propose a block circulant embedding technique and use Fast Fourier Transforms (FFT) for inference. By computing the Bayesian loss function tailored to switching Gaussian processes, an adaptive monitoring procedure is developed to select features from available sensors that optimize the trade-off between sensor power consumption and the prediction performance quantified by state prediction entropy. We demonstrate the effectiveness of our framework on the popular benchmark of UCI Human Activity Recognition using Smartphones.
\end{abstract}

\section{INTRODUCTION}

Smart health solutions are becoming ever more feasible with the rapid development of sensors and mobile applications that can continuously collect human behavioral data. Indeed, many sensors from various sources such as environmental \citep{poppe2007vision}, body-worn sensors \citep{lukowicz2004recognizing,karantonis2006implementation}, or even smartphone based sensors \citep{anguita2013public} are prevalent for health monitoring applications. With so many ambient sensors collecting different measurements, it becomes important not only to maintain good monitoring accuracy, but also low power consumption, to ensure effective and sustainable monitoring. Such a trade-off between monitoring accuracy and monitoring resource allocation is natural and ubiquitous in many applications \citep{he2006vigilnet,wiser2008annual,wu2018deep,wang2018energynet}. 

Such problems require an accurate characterization of the trade-off between monitoring resource usage and monitoring accuracy. In the case of human behavioral and health monitoring using ambient sensors, a power efficient sensing scheme can be achieved by deciding which group of sensors to use at a given time. Here, a trade-off arises between sensor energy usage and the uncertainty in ignoring certain sensor signals while monitoring. To characterize this trade-off, it is necessary to have a model that can quantify the uncertainty of the activity prediction with respect to the sensor measurements over time. With this model, a loss function associated with both energy cost and activity prediction uncertainty can be defined and therefore the adaptive monitoring problem can be solved by a sequential decision process.

Sequential decision process with Markov models have been well studied in the literature. For example, Markov decision process (MDP) and partially observed Markov decision process (POMDP) \citep{krishnamurthy2016partially} are popular stochastic models for sequential decision process. They can be applied for optimal preventive maintenance policy \citep{byon2010optimal} and path planning under uncertainty \citep{morere2017sequential}. Other decision process, for example Bayesian Optimization, have been applied for optimal sensor set selection \citep{garnett2010bayesian}. In this paper, activity detection is based on time series sensor signals, so the sensor selection problem depends on not only the current state, but also the history. The computation of the loss function will therefore be complicated due to the sequential decision nature.

Besides all above challenges in characterizing the resource limitation and monitoring accuracy trade-off, modeling the measurements taken from these ambient sensors is also challenging by itself. For example, smartphone sensor measurements are  abundant yet noisy, and thus require efficient computational methods to process effectively. Moreover, the sensor measurement frequency can vary with time, the duration of each activity may vary, and the multivariate effects between the large set of features can be difficult to capture. Methods to model such time-series measurements include linear dynamical systems~\citep{barber2012,ardywibowo2018}, ARMA models \citep{torres2005forecast}, Kalman filters \citep{harvey1990forecasting}, point processes \citep{gunawardana2011model}, and Recurrent Neural Networks \citep{funahashi1993approximation}. However, most of these existing methods focused on prediction only and did not attempt to characterize the uncertainty of the measurements.

To tackle this problem, we derive a switching multivariate Gaussian process model for the goal of activity recognition. Our model is a hierarchical one consisting of a Hidden semi-Markov model (HSMM) for the discrete activity states, and a multivariate Gaussian process to model both the time dependent and the inter-variable correlations between different sensor measurements. We use a block circulant embedding technique and use Fast Fourier Transforms (FFT) to speed up model inference and uncertainty quantification of our proposed model. Using this model, we then develop an adaptive monitoring scheme that for each monitoring period uses the optimal group of sensors by optimizing the Bayesian cost function considering both the activity predictive entropy\footnote{Here the predictive entropy is a measure for uncertainty, the reduction of which indicates the information provides by the selected sensors. It is computed approximately by Monte Carlo since it has no closed-form expression. } and the energy cost of selected sensors. 

We implement our model on the UCI Human Activity Recognition using Smartphones dataset \citep{anguita2013public}. This dataset consists of labeled trajectories of smartphone sensor measurements from multiple subjects under 6 different activities: walking, walking upstairs, walking downstairs, sitting, standing, and laying. The time series trajectories consist of features extracted from gyroscope and accelerometer measurements, such as movement angle, jerk, acceleration, and moving averages. Extensive results demonstrate the effectiveness of our framework on achieving competitive performance-energy trade-offs

\section{THE MODEL: SWITCHING GAUSSIAN PROCESS}

Throughout our presentation, we will use the following notation convention: bold faces indicate vectors or multivariate processes, capital letters indicate matrices or covariance functions of two variables, and script letters indicate sets.

\subsection{Model Formulation}

We first describe the model for time series sensor measurements from one subject. For the different activity types, we assume that there is an underlying semi-Markov jump process \citep{yu2010hidden} that governs the transitions between them. Denoting the discrete valued activity state at time $t$ as $x(t)$, the semi-Markov jump process can be expressed as:
\begin{equation} \label{eq1}
x(t) = \sum_{n=1}^{N}{x_n \textbf{1}_{\{ \tau_n \leq t < \tau_{n+1} \}}}.
\end{equation}
Here, $ x_n \in \{1, ..., A\} \equiv \mathcal{X} $ are discrete activity states that discriminate between the $ A $ different activity types, while $ \tau_n $ and $ \tau_{n+1} $ are the random start and end time points of the discrete activity state $ X_n $. We can denote the random duration that the subject stays in $ X_n $ as $ s_n = \tau_{n+1} - \tau_n $. We adopt an explicit-duration model for the sojourn times by modeling $ s_n $ using a Gamma distribution with state specific parameters as follows:
\begin{equation} \label{eq2}
s_n | (x_n = i) \sim Gamma(\bm{\gamma}_i), \forall i \in \mathcal{X},
\end{equation}
where $ \bm{\gamma}_i  = \{ k_i, \beta_i \} $ are the parameters for the duration distribution of activity $ i $ with $ k_i $ and $ \beta_i $ representing shape and scale parameters respectively. The advantage of this semi-Markov model is that it explicitly models the varying durations of different activity types. For example, in our application, we can see that the walking upstairs and walking downstairs activities all take less time to complete compared to the other activities. This activity duration may be informative in inferring the different activities. To complete the semi-Markov jump process model modeling the underlying activity states, the state transition probability themselves can be modeled by a transition probability matrix as follows:
\begin{equation} \label{eq3}
P(x_{n+1}=j | x_n = i) = p_{ij}.
\end{equation}

For the observation process $ \textbf{y}(t) = [y^1(t), ..., y^P(t)]^\top $, we incorporate a switching $ P $-variate Gaussian Process. This process models the dynamic heterogeneity of the time series measurements by switching between different Gaussian process models for each inferred activity state or context. Specifically, we model $ \bm{y}_n (t) $, the observation process occurring at time period $ [\tau_{n+1}, \tau_n] $, as follows:
\begin{equation} \label{eq4}
\textbf{y}_n(t) | (x_n = i) \sim \mathcal{GP}(\textbf{m}_i(t), \textbf{K}_i(t,t')).
\end{equation}
Here, conditioned on the activity state $ x_n=i $, the observation process $ \textbf{y}(t) $ will be a Gaussian process with activity-state-dependent mean and covariance function $ \textbf{m}_i(t) $ and $ \textbf{K}_i(t, t') $ respectively. The covariance function $ \textbf{K}_i(t, t') $ is a multivariate covariance function between two univariate variables $ y_n^a(t) $ and $ y_n^b(t) $ in $ \textbf{y}_n (t) $ for $a, b $ in the set of features $ \mathcal{Y} = \{1, \dots, P\} $.

With this, the observation process can be expressed as
\begin{equation} \label{eq5}
\textbf{y}(t) = \sum_{n=1}^{N}{\textbf{y}_n(t)\textbf{1}_{ \{ \tau_n \leq t < \tau_{n+1} \} }}.
\end{equation}

In order to handle the potential multivariate correlations between the sensor observations, we implement an intrinsic correlation model \citep{bonilla2008multi}. In this model, we assume that we can separate the covariance function of $ \textbf{y}_n(t) $ into a time dependent covariance function $ K_i^\mathcal{T} (t,t') $, and a free-form inter-variable covariance matrix $ K_i^\mathcal{Y} (a,b) $ between two variables $ y_n^a (t) $ and $ y_n^b (t) $ in $ \textbf{y}_n (t) $ as follows:
\begin{equation} \label{eq6}
K_i(t,t') = K_i^\mathcal{T}(t,t') \otimes K_i^\mathcal{Y}(a,b).
\end{equation}

Knowing the activity state and a set of noisy observations $ z $ at a set of time points $ S $, the prediction on a new set of points $ S^p $ for the $ l^{th} $ variable is given by:
\begin{equation} \label{eqPred1}
y_n^l(S^p) | (\textbf{z}, x_n = i) \sim N( m_i^l(S^p), \Sigma_i^l(S^p) ); 
\end{equation}
\begin{equation} \label{eqPred2}
m_i^l (S^p) = (\textbf{k}_i^{l,\mathcal{Y}} \otimes K_i^{(S, S^p), \mathcal{T}})^\intercal \Sigma_i^{-1} \textbf{z};
\end{equation}
\begin{equation} \label{eqPred3}
\Sigma_i = K_i^\mathcal{Y} \otimes K_i^\mathcal{T} + D \otimes I;
\end{equation}
\begin{eqnarray} \label{eqPred4}
\Sigma_i^l(t_p) & = & \textbf{k}_i^{l,\mathcal{Y}} \otimes K_i^{(S,S), T} - (\textbf{k}_i^{l,\mathcal{Y}} \otimes K_i^{(S, S^p), \mathcal{T}})^\intercal \times \nonumber \\
& & \Sigma_i^{-1} (\textbf{k}_i^{l,\mathcal{Y}} \otimes K_i^{(S, S^p), \mathcal{T}}).
\end{eqnarray}

\begin{figure}[!tb]
  \centering
  \includegraphics[width=0.5\textwidth]{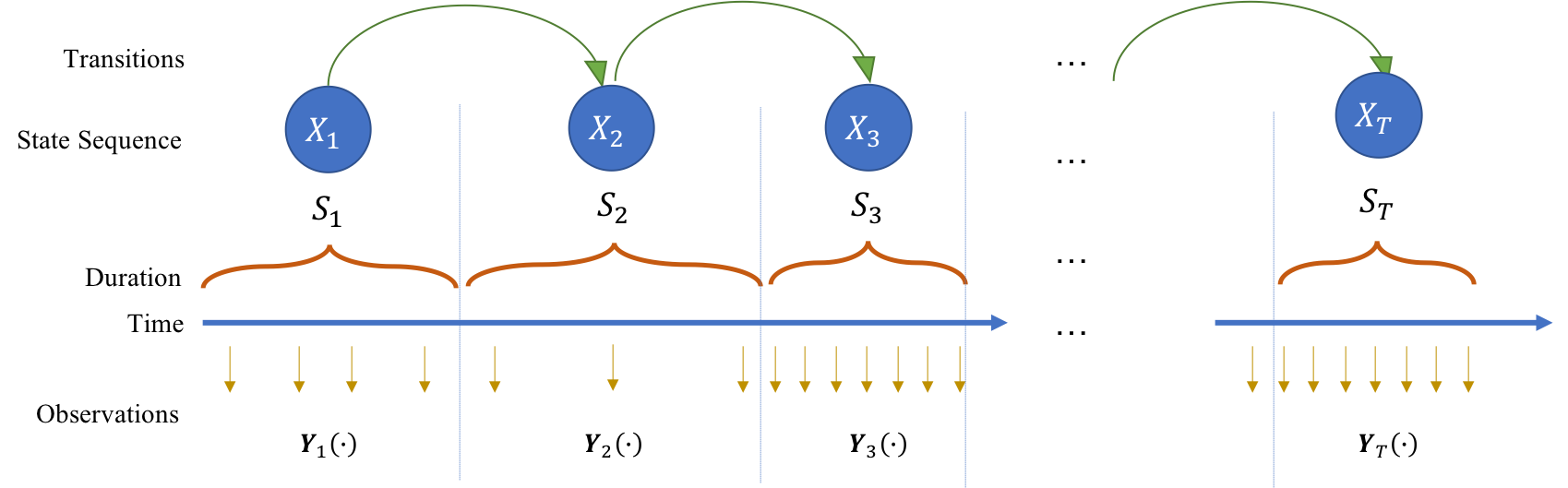}
  \caption{The hidden semi-Markov model.}
  \label{modelill}
\end{figure}

Here, $ \textbf{z} $ is a vector of noisy multivariate measurements structured as $ \textbf{z}=[z_1^1, z_2^1, ..., z_N^1, z_1^2, ...,z_N^P ]^T $, where $ z_t^p $ is the measurement of variable $ p $ at time $ t $. The vector $ \textbf{k}_i^{l,\mathcal{Y}} $ selects the $ l^{th} $ column of $ K_i^\mathcal{Y} $, while $ K_i^{(S,S^p),\mathcal{T}} $ is a matrix of time dependent covariances between the prediction time points and the observed time points. Finally, $ D $ is a $P \times P$ matrix of independent noise variances $ \sigma_p^2 $ for each multivariate variable $ p $.
The parameters of our model can be combined as: $ \bm{\theta} = \{ \bm{\gamma}_i, p_{ij}, K_i^\mathcal{T}(\cdot), K_i^\mathcal{Y}(\cdot), \forall, j \in \mathcal{X} \} $. Where the covariance matrices can be further parameterized. Specifically, we use a Matern covariance function for the time dependent covariance, while a Cholesky factorization was used for the multivariable covariance to ensure positive definiteness. An illustration summarizing the model formulation is shown in Figure \ref{modelill}.

\subsection{Model Inference}

We describe both the parameter inference and process inference algorithms for the proposed model with switching Gaussian process. For parameter inference, we use a population model to combine the data from multiple subjects, treating each subject as an independent realization of our model. We estimate the parameters of our model using maximum likelihood inference. Meanwhile, process inference can be efficiently done, similarly to classical filtering methods in traditional Hidden Markov Models (HMMs) \citep{rabiner1986introduction} using the Forward-Backward algorithm.

\subsubsection{Parameter Inference}

In our application, since all of the activity states are labeled in the time-series data, estimation of the semi-Markov jump process parameters is straightforward. Specifically, for each Gamma distributed activity state duration with parameter $ \bm{\gamma}_i=\{k_i, \beta_i \} $, we can derive the corresponding maximum likelihood estimates for $k_i$ and $\beta_i$ as follows:
\begin{equation} \label{eq7}
v_i = \log(\frac{1}{N_i} \sum_{n=1}^{N_i}{s_i^n}) - \frac{1}{N_i} \sum_{n=1}^{N_i}{s_i^n};
\end{equation}
\begin{equation} \label{eq8}
\hat{k}_i \approx \frac{3 - v_i + \sqrt[]{(v_i - 3)^2 + 24v_i}}{12v_i};
\end{equation}
\begin{equation} \label{eq9}
\hat{\beta}_i = \frac{1}{\hat{k}_i N_i} \sum_{n=1}^{N_i}{s_i^n},
\end{equation}
where $ s_i^n $ is the duration of the $n$-th time stamp in state $ i $. Besides, the transition probability can be inferred by simply counting the number of transitions between activity states as follows:
\begin{equation} \label{eqProb}
\hat{p}_{ij} = \frac{n_{ij}}{\sum_{j}{n_{ij}}},
\end{equation}
where $ n_{ij} $ is the number of transitions from state $ i $ to state $ j $.

To ensure positive definiteness of the multivariate covariance, we can parametrize it using the Cholesky decomposition $ K_i^\mathcal{Y}=LL^\top $ for lower-triangular matrix $ L $. Since each subject's process model is independent from each other given the model parameters, we can write the complete log-likelihood of all subjects as a sum of individual log-likelihoods. Specifically, denote by $ \mathcal{Q}_j $ as two times the negative log-likelihood of subject $ j $. This term can be expanded as follows: First, for each time frame $ \tau $ corresponding to a different activity state, we put our observations vector $ z $ in an $N_\tau \times P $ matrix form, denoted by $ Z_\tau $. Then, denote the matrix of corresponding mean functions for each $ Z_\tau $ as $ F_\tau $.
\begin{IEEEeqnarray}{rCl}
\mathcal{Q}_j &=& \sum_{\tau=1}^{T_j}{\bigg(N_\tau \log|K_{i_\tau}^Y| + P \log|K_{i_\tau}^T| + N_\tau P \log 2\pi} \nonumber\\
& &tr[(K_{i_\tau}^Y)^{-1} F_\tau^\intercal (K_{i_\tau}^T)] F_\tau] + N_\tau \sum_{p=1}^{P}{\log \sigma_p^2} +  \nonumber\\
& & tr[(Z_\tau - F_\tau) D^{-1} (Z_\tau - F_\tau)^\intercal ] \bigg),
\end{IEEEeqnarray}
where $ i_\tau $ is the activity state of subject $ j $ in time frame $ \tau $ with corresponding time dependent and multivariate covariance matrices $ K_{(i_\tau)}^\mathcal{T} $ and $ K_{(i_\tau)}^\mathcal{Y} $. With this, the complete log-likelihood $ \mathcal{Q} $ will be a sum of the individual log-likelihoods:
\begin{equation} \label{eq11}
-2\mathcal{Q} = \sum_{j=1}^{M}{\mathcal{Q}_j}.
\end{equation}

With this, any optimization method can be used to estimate the parameters of the multivariate Gaussian process model. The exact parameter updates are omitted from this presentation and the reader is referred to \citep{bonilla2008multi} for more details.

\paragraph{Fast Inference using Block Circulant Embedding}
Computing the likelihood as well as the gradient of the multivariate Gaussian Process models can be computationally challenging when dealing with a large number of features and the potential interactions between them. This is mainly due to the inversion and determinant calculation operations that need to be performed on the covariance matrix $ \textbf{K}_i(t,t') $. 

To speed up these computations, we propose a block circulant embedding approach. We first note that the measurements of our dataset are evenly spaced in time and assume that we only have one feature measurement. For this setting, the covariance matrix can be expressed as follows:
\[
C =
\begin{bmatrix}
    C(0)   & C(1)   & C(2)   & \dots  & C(T)   \\
    C(1)   & C(0)   & C(1)   & \dots  & C(T-1) \\
    \vdots & \vdots & \vdots & \ddots & \vdots \\
    C(T)   & C(T-1) & C(T-2) & \dots  & C(0)
\end{bmatrix}.
\]
\normalsize
Here $ C(t) $ are scalars as we only have a single feature. Observe that we can embed this matrix in the following larger circulant matrix: $\tilde{C}=$
\small
\[
\begin{bmatrix}
    C(0)   & \dots  & C(T)   & C(T-1) & \dots  & C(1)   \\
    \vdots & \ddots & \vdots & \vdots & \ddots & \vdots \\
    C(T)   & \dots  & C(0)   & C(1)   & \dots  & C(T-1) \\
    C(T-1) & \dots  & C(1)   & C(0)   & \dots  & C(T-2) \\
    \vdots & \ddots & \vdots & \vdots & \ddots & \vdots \\
    C(1)   & \dots  & C(T-1) & C(T-2) & \dots  & C(0)   \\
\end{bmatrix},
\]
\normalsize
which is fully specified by the first row vector $ \textbf{c} = [C(0), C(1), \dots C(T), C(T-1), \dots C(1) ] $. We can perform an eigenvalue decomposition of the above matrix as follows:
\begin{equation} \label{eqEmbedding}
\tilde{C} = F \Lambda F^\intercal.
\end{equation}
$F^\intercal $ and $ F $ denote the Fourier and inverse Fourier transform respectively, and $ \Lambda = F^\intercal \textbf{c} $. Using the Fast Fourier Transform (FFT), this calculation can be done in $ O(T \log(T)) $. Consequently, inverse and determinant calculations can be done at this same time complexity \citep{davis2012circulant}. This idea can be extended to the case of block circulant matrices. First we note that the covariance matrix $ K_i (t, t') $ can be embedded into a block circulant matrix specified by the matrix $ \textbf{K}_i = [K_i^\mathcal{T}(0), \dots K_i^\mathcal{T}(T), K_i^\mathcal{T}(T-1), \dots K_i^\mathcal{T}(1) ] \bigotimes K_i^\mathcal{Y} $. We then notice that each vector $ [K_i^\mathcal{T}(0), \dots K_i^\mathcal{T}(T), K_i^\mathcal{T}(T-1) \dots K_i^\mathcal{T}(1)] K_i^\mathcal{Y}(a, b), \forall a,b \in \{1, \dots P\} $ defines a circulant submatrix. To compute the determinant and eigen-decomposition of each block circulant matrix, it suffices to simply decompose each of them using FFT. Finally, the model likelihood can be computed in $ O(P^2 T \log(T)) $.
\begin{figure*} 
  \centering
  \includegraphics[width=1\textwidth]{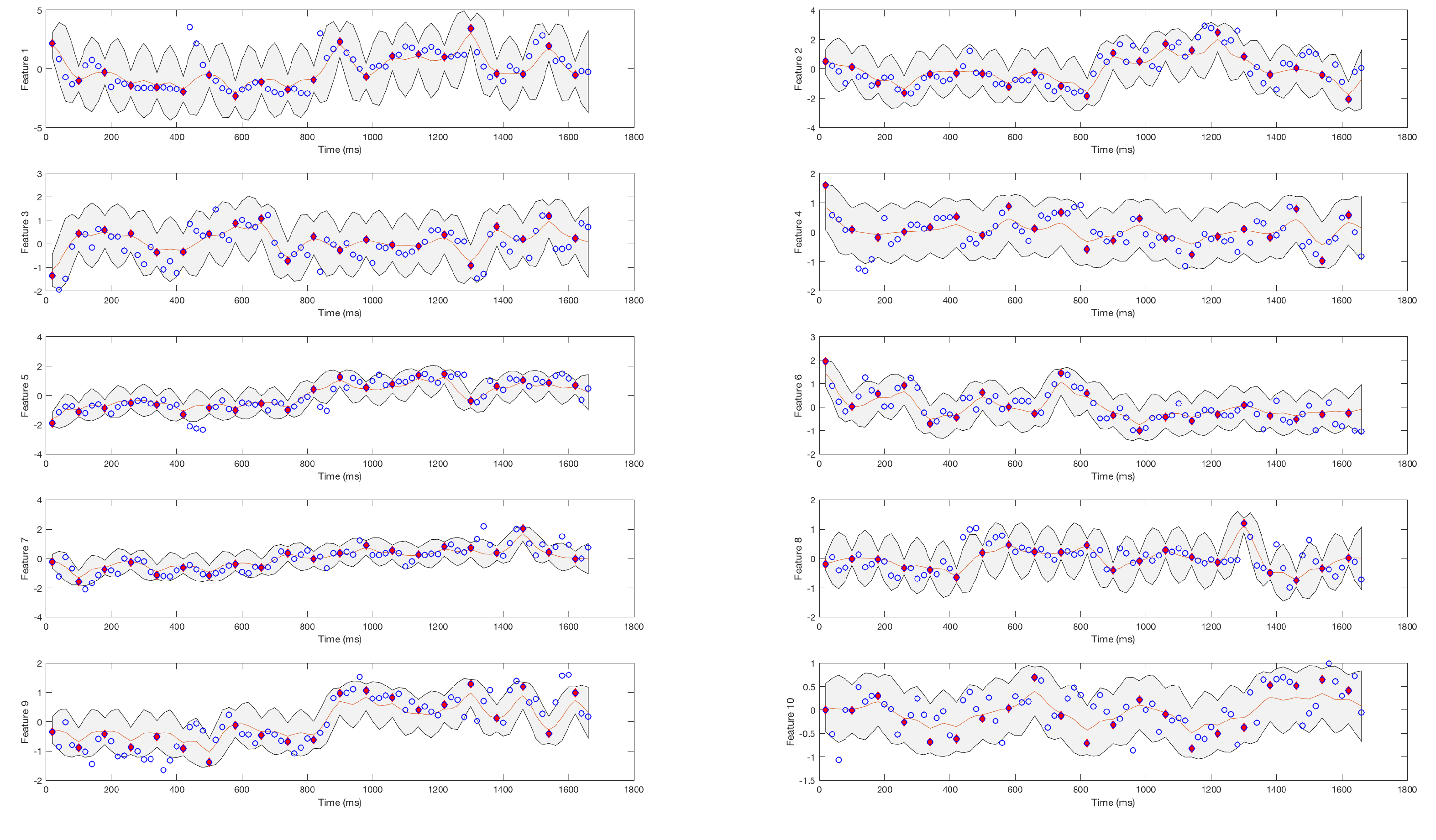}
  \vspace{-2.5em}
  \caption{Prediction trajectory comparison of different features.}
  \label{multivarpred}
\end{figure*}

\subsubsection{Process Inference}

Activity state inference can be conducted using standard filtering approaches similar to classical Hidden Markov Models (HMM) \citep{rabiner1986introduction}. Specifically, we are interested in estimating the current state at time $ t $ given all previous observations. 

Firstly, denote $ y_\tau(t_0:t_1 ) $ as a collection of variables $ y_\tau $ for $ t_0 \leq \tau \leq t_1 $. The expression $ x[t_1:t_2 ] = i $ indicates that $ x $ is in state $ i $ between times $ t_1 $ and $ t_2 $ but no later or sooner. In other words,  $ x $ will be in a different state before and after time $ t_1 $ and $ t_2 $ respectively. With this, the activity detection problem can be expressed by the following maximum \emph{a posteriori} (MAP) filtering problem:
\begin{equation} \label{eq12}
\hat{x}_t = argmax_{i \in \mathcal{X}} P(x_t = i, \textbf{y}_{t_0 : t} | \bm{\theta}).
\end{equation}
It can be solved recursively using dynamic programming. Specifically, at any given time point we need to compute the following forward and backward variables respectively:
\begin{equation} \label{eq13}
\alpha_t (j, d) \coloneqq P(x_{[t-d+1 : t]} = j, \textbf{y}_{1:t} | \bm{\theta});
\end{equation}
\begin{equation} \label{eq14}
\beta_t(j,d) \coloneqq P(\textbf{y}_{t+1 : T} | x_{[t-d+1 : t]} = j, \bm{\theta}).
\end{equation}

Since we are merely interested in filtering the current activity states or contexts in an online fashion, we need only compute the forward variables $ \alpha_t(j,d) $. The recursion for computing them can be expressed as follows:
\begin{eqnarray} \label{eq15}
&&\alpha_t (j, d) \\ &=& \sum_{i \in \mathcal{X} \backslash \{ j \}}{\sum_{d' \in \mathcal{D}}{ \alpha_{t-d}(j, d') \cdot a_{(i,d')(j,d)} b_j(\textbf{y}_{t-d+1:t} ) } }. \nonumber
\end{eqnarray}

Here, $ a_{(i,d')(j,d)} $  is the transition probability from staying in state $ i $ for duration, $ d' $, towards staying in state $ j $ for duration $ d $. Based on our model, this is a semi-Markov jump process, with state specific durations modeled as gamma distributions. Hence, this probability can be expressed as:
\begin{IEEEeqnarray}{rCl}
a_{(i,d')(j,d)} &\coloneqq& P( x_{[t-d+1:t]} = j | x_{[t-d-d'+1 : t-d]} = i ) \nonumber\\
                & = & Gamma(d' | \gamma_i) P(x_{t-d+1} = j | x_{t-d} = j) \times \nonumber\\ 
                &   & Gamma(d | \gamma_j) \nonumber\\
                & = & Gamma (d' | \gamma_j) P_{ij} Gamma(d | \gamma_j).
\end{IEEEeqnarray}

In \eqref{eq15}, $ b_{(j,d)} (\textbf{y}_{(t-d+1:t)} ) $ denotes the observation probability that is modeled as a switching Gaussian process. Specifically, this quantity can be expressed as follows:
\begin{IEEEeqnarray}{rCl}\label{eq17}
b_j(\textbf{y}_{t+1 : t+d}) & \coloneqq & P(\textbf{y}_{t+1 : t+d} | x_{[t+1 : t+d]} = j) \\
                   & = & N(\textbf{y}_{t+1 : t+d} | \textbf{m}_j^{t+1 : t+d}, K_j^{t+1 : t+d}), \nonumber
\end{IEEEeqnarray}
where $ \textbf{m}_j^{(t+1:t+d)} $ and $ K_j^{(t+1:t+d)} $ are the predicted means and covariances from time points $ t+1$ to $t+d $.

\section{THE ADAPTIVE MONITORING FRAMEWORK}
In this section, an adaptive monitoring method is proposed based on the hidden semi-Markov model with switching Gaussian Process. In particular, with the previous process inference procedure, we can derive the activity state predictive distribution. The prediction uncertainty can then be characterized by the entropy of the derived distribution. In the cases when we are confident about the activity state in the next time point based on the predictive entropy, we may not need all the sensor data for the next observations, and therefore we can choose only a subset of the sensors for the sake of saving energy. Adaptive monitoring will determine which group of sensor measurements may not be necessary, if we predefine groups of features based on the senor monitoring resource allocation requirements. 


The set of predefined feature groups can be denoted as $ \mathcal{F} $, from which one of the feature group $m\in \mathcal{F} $ is chosen at each observation time $t$ to observe signals $\textbf{y}_t^{m}$. For each feature group $m$, we also define its cost related to power consumption, which is set to a constant $\lambda_m $ for simplicity. The selected feature group should balance the energy cost and prediction uncertainty gain. This is achieved by minimizing the following loss over all predefined feature groups:
\begin{equation}\label{eq18}
	\mathcal{L}(m) = \mathbb{E}_{P(\textbf{y}_{t+1}^m|\textbf{y}_{1:t}^\mathcal{F})}[H[P(x_{t+1}|\textbf{y}_{1:t}^\mathcal{F}, \textbf{y}_{t+1}^m)]]+\lambda_m, 
\end{equation}
where $\textbf{y}_{1:t}^\mathcal{F}$ denotes all the previous observations of feature groups selected from $\mathcal{F}$ following the same rule, 
$H[p(x)] = -\sum p(x)\log p(x)$ is the entropy function. In the loss function, the first term averages over the predictive distribution of $\textbf{y}_{t+1}^m$ given $\bm{y}_{1:t}^\mathcal{F}$ and it describes the remaining uncertainty of $x_{t+1}$ after observing $\textbf{y}_{t+1}^m$, we would like to minimize this value to make sure that ignoring feature group $m$ does not increase much prediction uncertainty. At the same time, we also would like to achieve small energy cost $\lambda_m$.

\begin{figure*}[!ht]
  \centering
  \includegraphics[width=1\textwidth]{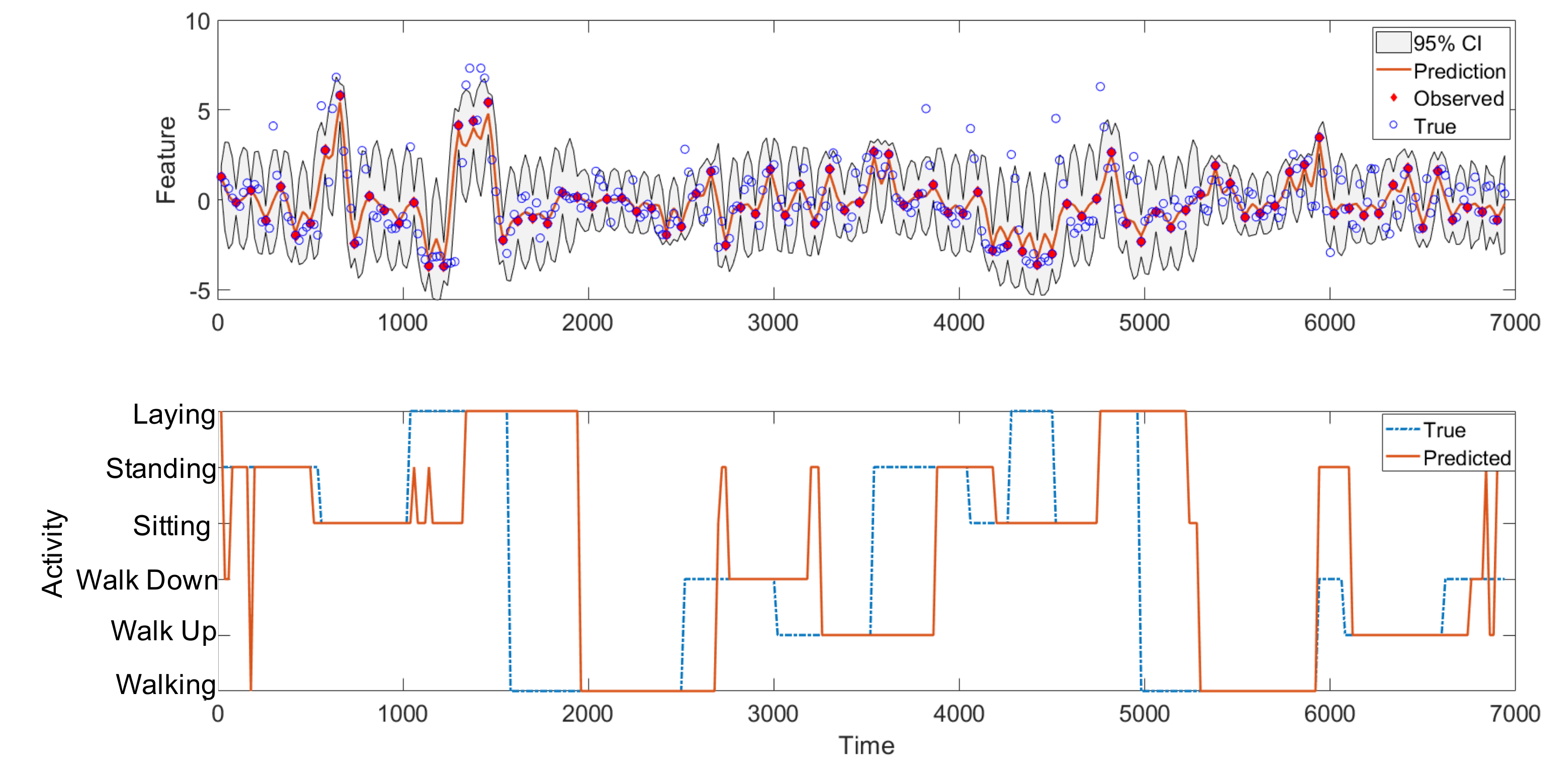}
  \vspace{-2em}
  \caption{Prediction trajectory for activity recognition and prediction.}
  \vspace{-0.5em}
  \label{fullpred}
\end{figure*}

\paragraph{Loss Function Computation}
The minimization of~\eqref{eq18} does not have an analytic solution form and must be solved approximately. We first approximate the expected predictive entropy in~\eqref{eq18} by drawing samples $\bm{y}_{t+1}^{m(i)}$ from $P(\bm{y}_{t+1}^m|\bm{y}_{1:t}^\mathcal{F})$. For each of these samples, we then calculate the corresponding entropy function $H[P(x_{t+1}|\textbf{y}_{1:t}^\mathcal{F}, \textbf{y}_{t+1}^{m(i)})]$. 

To derive the sampling probability $P(\textbf{y}_{t+1}^m|\textbf{y}_{1:t}^\mathcal{F})$, we adopt the following message passing algorithm:
\begin{eqnarray}\label{eq19}&&P(\textbf{y}_{t+1}^m|\textbf{y}_{1:t}^\mathcal{F}) \\&=& \sum_{i, j \in \mathcal{X}}{\sum_{d', d \in D}{b_j(\textbf{y}_{t+1}^m)} a_{(i,d')(j,d)}}\alpha_{t-d+1}(j,d'), \nonumber
\end{eqnarray}
which gives a Gaussian mixture distribution with respect to $\bm{y}^m_{t+1}$.
For each samples $\textbf{y}_{t+1}^{m(i)}$, we can easily calculate $P(\textbf{y}_{t+1}^m|\textbf{y}_{1:t}^\mathcal{F}, \textbf{y}_{t+1}^{m(i)})$ based on~(\ref{eq13}). 
Then the loss function can be approximated by:
\begin{equation}
	\mathcal{L}(m) = \lambda_m +\frac{1}{N}\sum_{i = 1}^{N}H[P(x_{t+1}|\textbf{y}_{1:t}^\mathcal{F}, \textbf{y}_{t+1}^{m(i)})],
\end{equation}
where $N$ is the number of samples drawn from $P(\textbf{y}_{t+1}^m|\textbf{y}_{1:t}^\mathcal{F})$.
\section{RESULTS AND DISCUSSION}
\begin{figure}[!ht] 
  \centering
  \includegraphics[width=0.5\textwidth]{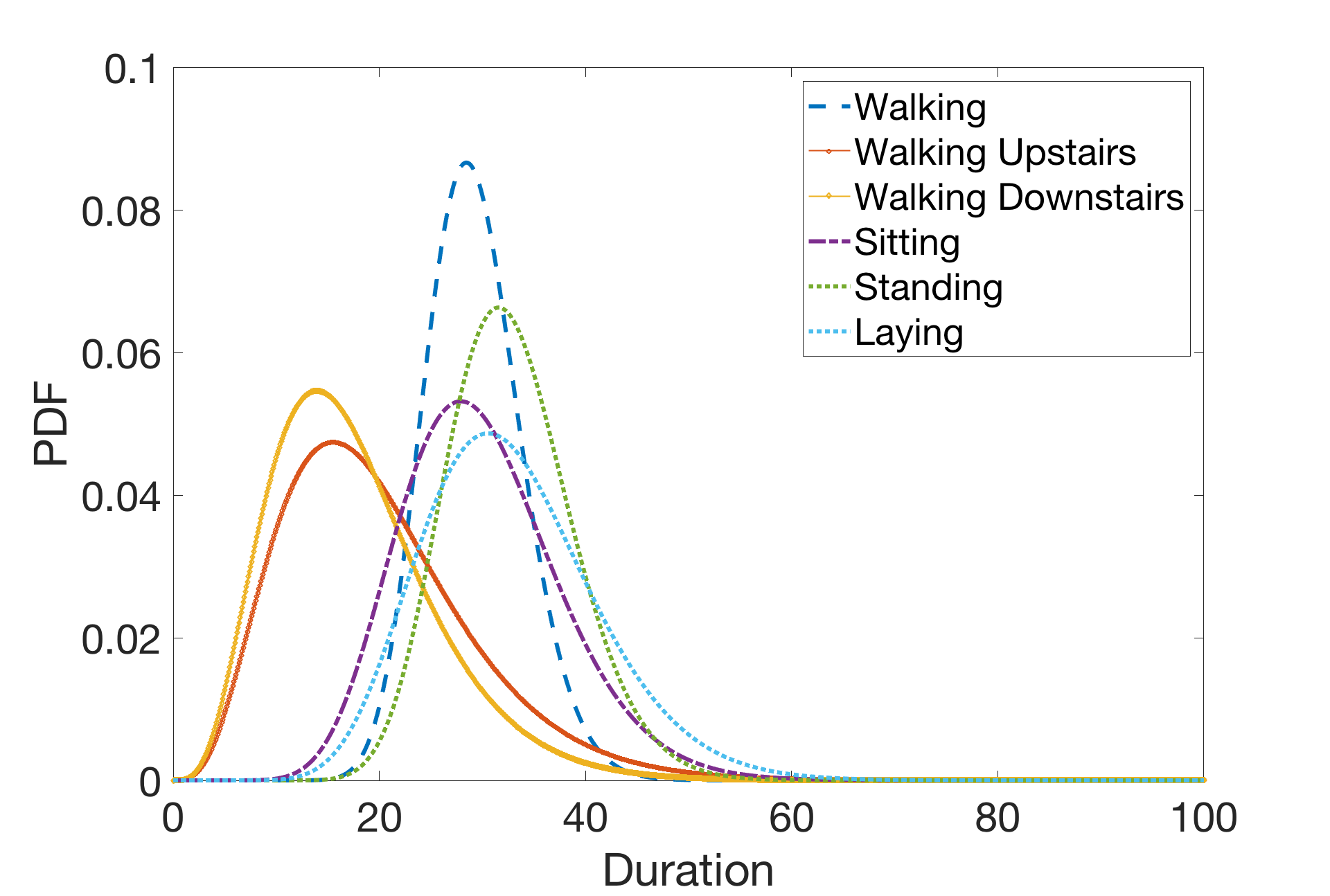}
     \vspace{-1em}
  \caption{Learned Gamma duration distributions of the different activity types}
  \label{gampdf}
\end{figure}
We apply our model to the popular UCI benchmark of ``Human Activity Recognition using Smartphones'' \citep{anguita2013public}. This dataset contains sensor measurements from a group of 30 volunteers performing six types of activities of daily living, including ``walking'' (1), ``walking upstairs'' (2), ``walking downstairs'' (3), ``sitting'' (4), ``standing'' (5), and ``laying'' (6). The sensor measurements are all taken at a constant rate and labeled manually with the corresponding activity type. We pre-process the data by Principal Component Analysis (PCA) to reduce computation overhead. Specifically, 10 principal components are derived from the 561 original features. We assume a constant mean for each activity by computing the population mean corresponding of each.

We evaluate the performance through two sets of evaluations. For the first evaluation, we assume activity states are known, and evaluate by the mean squared error (MSE) and absolute difference error (ABS) for trajectory prediction. In our second evaluation, we conduct joint activity recognition and trajectory prediction, where we only have observations up to the time of prediction. Several different Gaussian process setups and assumptions are compared to select the best performing model.

\begin{table}[!ht] 
\centering
\caption{Prediction result comparison at different Gaussian process setups.}
\begin{tabular}{p{50mm} c c}
\hlineB{3}
Model Setup      & MSE & ABS \\ 
\hlineB{3}
Baseline + Separate time dependence + Separate Multivariate & 0.4988 & 0.4480 \\ 
\hline
Baseline + Separate time dependence + Combined Multivariate & 0.4040 & 0.4244 \\ 
\hline
Separate time dependence + Combined Multivariate            & \textbf{0.3852} & \textbf{0.4235} \\ 
\hline
\end{tabular}
\label{acctable1}
\end{table}

\begin{figure}[!ht] 
  \centering
  \includegraphics[width=0.485\textwidth]{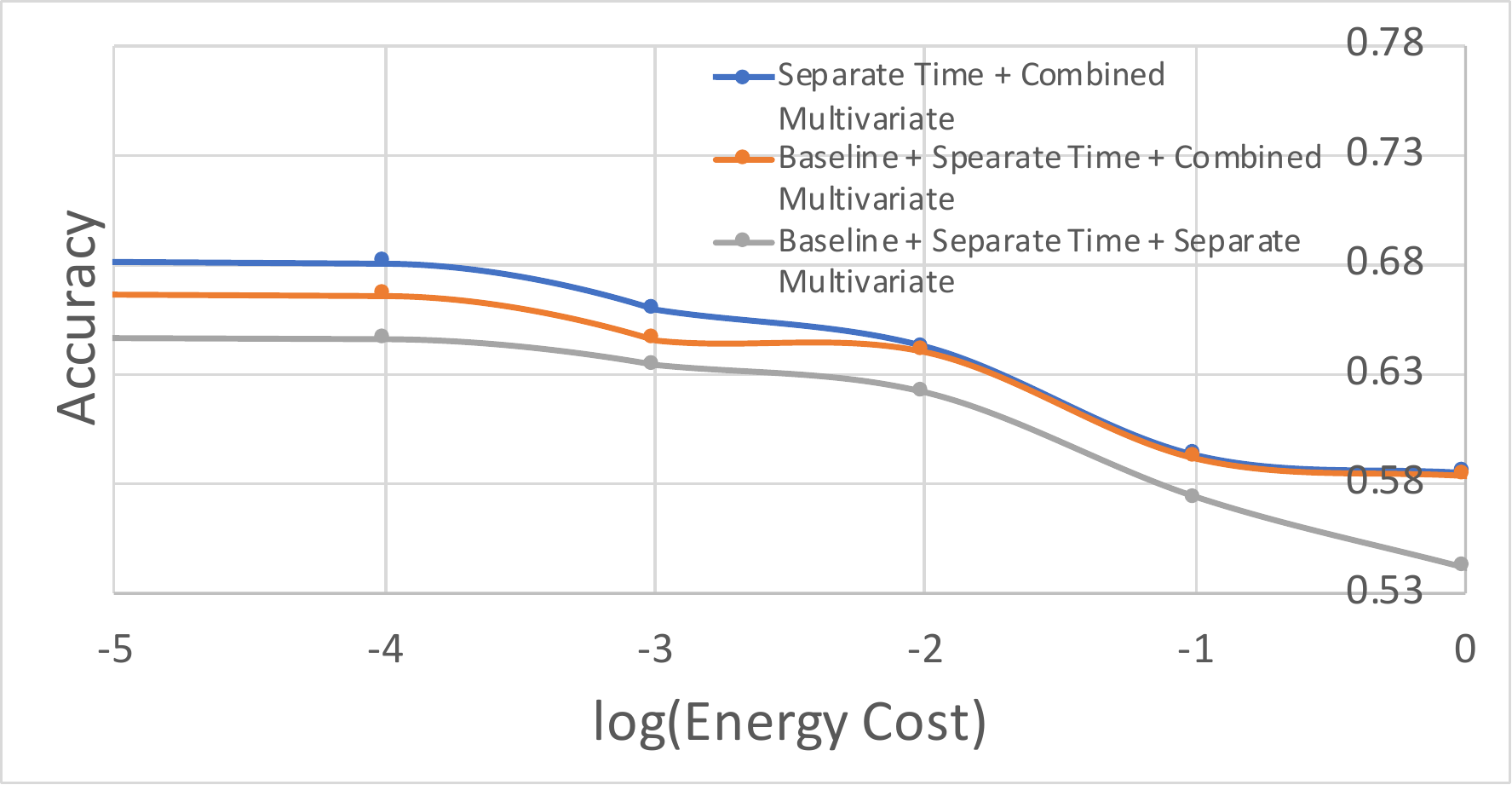}
    \vspace{-1em}
  \caption{Energy cost and accuracy trade-off curve for activity recognition}
  \label{energycost}
\end{figure}

\subsection{Trajectory Performance Prediction assuming Known Activity States}

\begin{figure*}[!ht] 
  \centering
  \includegraphics[width=0.935\textwidth]{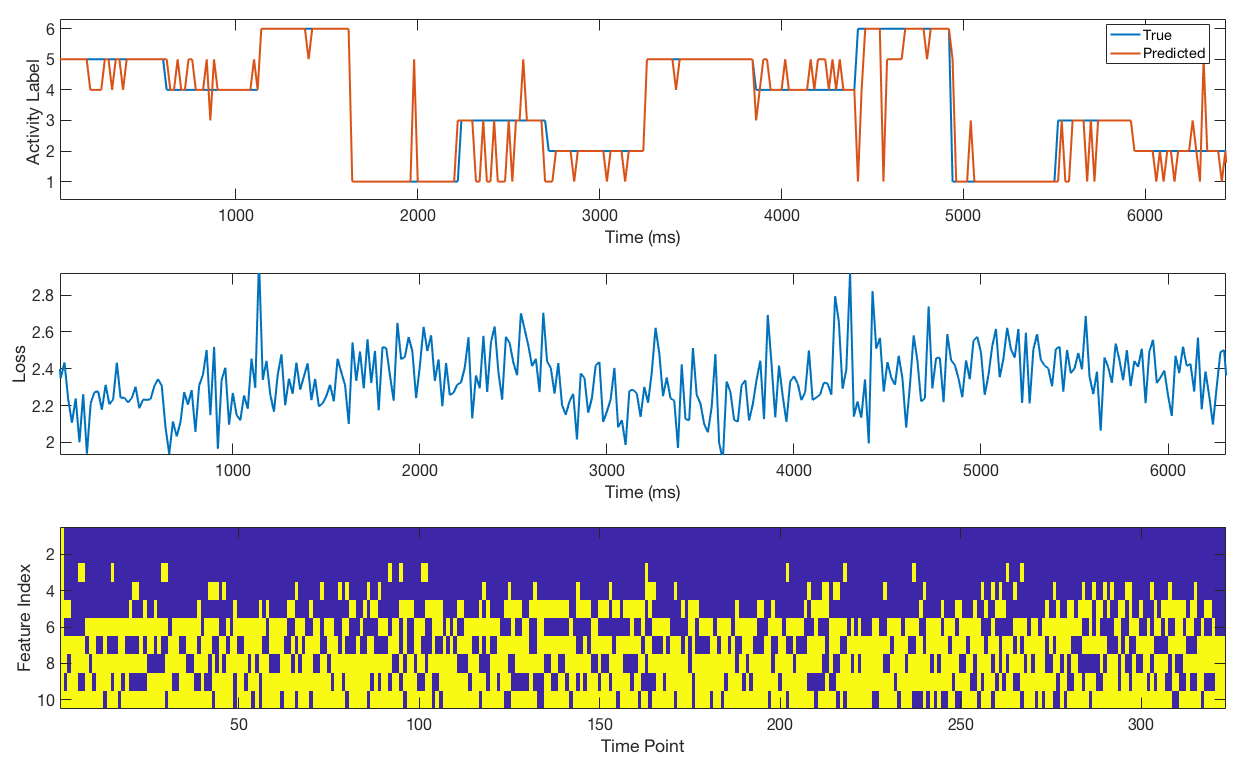}
    \vspace{-1em}
  \caption{Context prediction results, loss function values, and active features across time for $\lambda = 0.1$.  Yellow highlights indicate active sensors.}
  \label{energycostpred}
\end{figure*}

We compare the sensor signal trajectory prediction performance under three different Gaussian process model assumptions. The first assumption uses two Gaussian process components: a baseline model to model the entire trajectory of all patients and an activity specific model to capture the varying dynamics of the different activities. We assume that either the temporal and multivariate dependencies are modeled separately for each activity, or the multivariate dependencies are combined between different activities. The prediction accuracy of these two models are shown in the first two entries of Table~\ref{acctable1}. Our best performing model, however, considers only a single multivariate effect model for different activities with no special consideration for any baseline trajectory. We use this best performer's assumption for the remaining experiments. We set the ratio between observed measurements and held-out measurements to be 1:4. Figure \ref{multivarpred} demonstrates prediction results of different variables where the model is able to follow the trajectories of the held-out data accurately. Moreover, the sensor signals for different activity states also have different variances, indicated by varying widths of the predicted confidence intervals. 


We compare the activity state prediction results in two tasks. Figure \ref{gampdf} illustrates the first task, where the learned semi-Markov jump process duration distributions manifest a difference in duration between walking upstairs and downstairs compared to the other activities. Figure \ref{fullpred} demonstrates activity state prediction results in our second task. In this task, the average testing prediction accuracy can go up to 74.21\%. Additionally, a large portion of inaccurate estimations are due to the time lag between the actual and predicted activity switch, which is a common problem in filtering tasks that need a certain amount of measurements in order to be confident in an activity switch.

\subsection{Adaptive Monitoring}
   \vspace{-1em}
   
Finally, We implement the adaptive monitoring on the same UCI dataset with same training and testing setup. To perform sensor selection, we treat the 10 principal components as 10 different sensors and assume they have the same energy cost. We then define feature subsets consisting of all possible combinations of 4, 7, and 10 sensor measurements. Thus, the energy cost for each group is proportional to the number of measurements in the group. In our experiments, we vary the energy cost $ \lambda $ of a single sensor from 0 to 1 and view its impact on prediction performance.

As shown in Figure \ref{energycost}, we see that as the energy cost $ \lambda $ is increased, the prediction accuracy decreases as each sensor measurement comes at a higher cost. On the other hand, when there is no energy cost, the monitoring plan will prefer to utilize all of the sensor measurements available to perform the best prediction possible. A sample of the context prediction trajectory with energy cost $\lambda = 0.1$, which exhibits the most activity in adaptive feature selection, is shown in Figure \ref{energycostpred}. In this specific example, we are able to get an accuracy of 79.26\% with an average sensor usage of 73.42\%. The most used features are the leading principal components indexed starting from 10. This indicates that the leading principal components are better correlated to the activity recognition task compared to lower principal components. Moreover, this also shows that our adaptive monitoring framework can detect the most relevant principal component features without knowing their order in advance.

\section{CONCLUSION}
   \vspace{-0.5em}
In this paper, we propose a hierarchical model consisting of a multivariate switching Gaussian process to model the signals based on different activity types. We applied our model on trajectory and activity prediction with the UCI dataset for model verification. MSE for trajectory prediction can be as small as 0.3852, and activity recognition accuracy can reach 74.21\%. Based on this model, we proposed an adaptive monitoring approach balancing the trade-off between sensor energy cost and prediction uncertainty. Within this monitoring scheme, we 
characterize the trade-off between monitoring accuracy and sensor energy efficiency. We show that our activity recognition scheme can stay robust and perform well under energy restrictions.

   \vspace{-0.3em}
\paragraph{ACKNOWLEDGEMENT} This project is in part supported by the Defense Advanced Research Projects Agency (FA8750-18-2-0027) and the National Science Foundation Awards CCF-1715027 and CCF-1718513. Part of the computing time is provided by the Texas A\&M High Performance Research Computing.

\bibliography{reference}
\bibliographystyle{aaai}

\end{document}